\documentclass{article}
\usepackage{ijcai24}

\usepackage{float}
\usepackage{times}
\usepackage{multicol}
\usepackage{multirow}
\usepackage{soul}
\usepackage{url}
\usepackage[hidelinks]{hyperref}
\usepackage[utf8]{inputenc}
\usepackage[small]{caption}
\usepackage{graphicx}
\usepackage{amsmath}
\usepackage{amsthm}
\usepackage{natbib}
\usepackage{booktabs}
\usepackage{algorithm}
\usepackage{algorithmic}
\newtheorem{definition}{Definition}

\usepackage[switch]{lineno}

\newtheorem{example}{Example}

\title{Formal Abductive Explanations for Navigating Mental Health Help-Seeking and Diversity in Tech Workplaces}

\author{
Belona Sonna$^1$
\and
Alain Momo$^2$
\and Alban Grastien$^1$\\
\affiliations
$^1$Australian National University\\
$^2$Australian Reinforcing Company\\
\emails
belona.sonna@anu.edu.au,
momoalain0505@gmail.com,
alban.grastien@anu.edu.au
}

\begin{document}

\maketitle

\begin{abstract}
This work proposes a formal abductive explanation framework designed to systematically uncover rationales underlying AI predictions of mental health help-seeking within tech workplace settings. By computing rigorous justifications for model outputs, this approach enables principled selection of models tailored to distinct psychiatric profiles and underpins ethically robust recourse planning. Beyond moving past ad-hoc interpretability, we explicitly examine the influence of sensitive attributes such as gender on model decisions, a critical component for fairness assessments. In doing so, it aligns explanatory insights with the complex landscape of workplace mental health, ultimately supporting trustworthy deployment and targeted interventions
\end{abstract}
\section{Introduction}

The global burden of mental health disorders is immense, affecting over 970 million people worldwide and posing substantial social and economic challenges \citep{Gupta2024AI}. In workplace contexts, mental health concerns ranging from depression and anxiety to stress-related disorders are particularly prevalent, yet often underreported and undertreated due to stigma or lack of structural support \citep{Jin2023AI}. As organizations increasingly turn to AI to predict, monitor, or support mental health interventions, critical questions arise about the appropriateness and trustworthiness of such systems \citep{Ali2024Ethics, Taylor2025Expectations}.

Recent studies underscore the promise of AI models in detecting early signs of psychiatric distress \citep{Gupta2024AI}, enhancing access in resource-limited settings \citep{Kumar2024GlobalMedi}, and tailoring interventions \citep{Doraiswamy2019Survey}. However, these benefits are tempered by concerns over algorithmic bias, lack of transparency, and the mismatch between predictive patterns and clinical reality \citep{Stefanis2024Trends, Zhu2024Risks}. Importantly, mental health is not a uniform concept but a heterogeneous collection of conditions, each influenced by distinct psychosocial and occupational factors \citep{Omar2024Bias}.

This heterogeneity means that the same AI model may rely on very different reasons to predict help-seeking across individuals. Thus, there is a pressing need for methodologies that do not merely interpret model outputs but formally justify why a particular model is suitable (or unsuitable) for specific mental health contexts, ensuring ethically sound, clinically relevant, and personalized care pathways.

\section{Problem Statement and Hypothesis}
Despite the growing use of AI in predicting mental health outcomes in workplace settings, there remains a critical gap in formally understanding why models make specific treatment-related predictions, particularly given the diverse manifestations of mental health conditions. Existing interpretability approaches often yield fragmented or heuristic insights, which are insufficient for assessing whether an AI model’s reasoning genuinely aligns with the multifaceted nature of psychiatric disorders. This lack of structured, auditable justification complicates efforts to ensure ethical deployment, appropriate intervention planning, and sustained clinician trust

We hypothesize that the use of formal abductive explanations can help bridge this gap by systematically uncovering minimal, context-specific reasons that drive model predictions. This, in turn, will reveal the heterogeneity of factors influencing help-seeking across different mental health contexts and enable principled selection of AI models best suited to distinct psychiatric needs, thereby enhancing both transparency and clinical relevance. 

Although this work is preliminary, we conducted an empirical study on the \textit{Mental Health in Tech Survey} dataset to establish the following contributions:
\begin{enumerate}
    \item Train a neural network as a baseline predictive model for help-seeking behavior.
    \item Compute formal abductive explanations to systematically uncover the reasons behind each decision of the trained network.
    \item Leverage these explanations to assess potential bias, particularly with respect to sensitive attribute such as gender.
    \item Analyze the impact of each feature on the outcome of the neural network.
    \item Identify the most prominent combinations of features that are likely to trigger predictions of seeking treatment or not.
\end{enumerate}
\section{Preliminaries}
To ensure clarity, we first introduce the formal notation, the dataset and the decision process used throughout this paper.
\subsection{Notation}
\begin{itemize}
  \item $V = \{v_1, \dots, v_n\}$ denotes a set of $n$ Boolean variables (features).
  \item An \emph{individual} $x = [x_1, \dots, x_n]$ is represented as a vector of Boolean values corresponding to the features in $V$. We interpret $x$ both as a vector and as a conjunction (or set) of literals, where each $x[i]$ is either $v_i$ or $\neg v_i$.
  \item $X$ denotes the space of all possible individuals.
  \item $D$ is the domain of the decision function, with $d \in D$.
  \item $\Delta: X \rightarrow D$ is the \emph{decision function} computed by the model.
  \item $XP$ is a \emph{property} of an individual $x$, defined as a subset (i.e., conjunction) of literals from $x$, such that $XP \subseteq x$. We denote by $vars(XP)$ the set of features that appear in $XP$.
  \item $p \in V$ denotes the \emph{protected attribute}.
  \item $X_{(V')}^x$ denotes the set of individuals that are identical to $x$ except (potentially) on the subset of features $V' \subseteq V$.
\end{itemize}
\subsection{Dataset}
We leverage the \textit{Mental Health in Tech Survey} dataset, which captures demographic, occupational, and mental health-related attributes of individuals in the tech industry, along with self-reported treatment-seeking behavior. The dataset comprises 1,257 instances, with 78.84\% identifying as male and 21.16\% as female or transgender. Among the respondents, 635 reported having sought treatment, while 622 did not.
For simplicity, we binarized the dataset ($X$) according to the procedure described in Appendix~\ref{Apx::A1}.

\subsection{The Decision process}
 In this paper, We have trained a feedforward neural network with slkearn-library to predict whether an individual seeks mental health treatment. The resulting model achieves an accuracy of \textbf{74\%}, with additional details and the full classification report provided in Appendix~\ref{Apx::A2}. Then the trained model has been converted as a SMT model following the listed steps:
 \begin{itemize}
     \item Extract the model's weights and structure: input weights matrix, bias vector, activation function and the output layer.
     \item Express each neuron as a linear formula using the weights and the bias term
     \item Write similar formulas for all hidden neurons and then for the output neuron.
     \item Combine them into a full logical expression that will be the decision function ($\Delta$) of the model.
 \end{itemize}
\section{What is formal Abductive explanations?}
Formal abductive explanations (AXPs) have been extensively studied in works such as \cite{Darwiche2020, Huang2023FromRT, Izza2023DeliveringIE, MarquesSilva2022LogicBasedEI, MarquesSilva2023DisprovingXM, Marques-Silva_Ignatiev_2022}, where an AXP is defined as follows:
\begin{definition}
    A formal abductive explanation (AXP) is a minimal set of features of an individual that is sufficient to guarantee the same decision by the model. Any individual sharing identical values on this minimal set of features will necessarily receive the same prediction.
\end{definition}
Algorithm~\ref{algo::CEalgorithm} computes a minimal abductive explanation for an individual $x$. The algorithm operates greedily: it attempts to remove each literal from $x$ and retains only those whose removal changes the outcome, thereby preserving minimal sufficiency.
\begin{algorithm}[h]
    \caption{Compute-Explanation}
    \label{algo::CEalgorithm}
    \textbf{Input}: Decision process $\Delta$\\
    \textbf{Input}: Individual $x$ \\
    \textbf{Input}: Decision $d = \Delta(x)$ \\
    \textbf{Output}: $XP$, a minimal abductive explanation for $x$
    
    \begin{algorithmic}[1]
        \STATE $XP \leftarrow x$
        \FOR{each feature $v \in V$}
            \STATE $\ell \leftarrow x[v]$
            \STATE $XP \leftarrow XP \setminus \{\ell\}$
            \IF{$\exists x' \supseteq XP$ such that $\Delta(x') \neq d$}
                \STATE $XP \leftarrow XP \cup \{\ell\}$
            \ENDIF
        \ENDFOR
        \RETURN $XP$
    \end{algorithmic}
\end{algorithm}
\begin{example}
    Let us compute a minimal abductive explanation for the decision of an individual represented by the feature vector
    \[
    x=[1, 1, 0, 1, 1, 0, 1, 1, 1, 0, 0, 0, 0, 0, 0, 1, 1, 0, 0],
    \]
\end{example}
\begin{table*}[t]
  \centering
  \caption{Computing an explanation for individual $[1, 1, 0, 1, 1, 0, 1, 1, 1, 0,0,0, 0, 0, 0, 1, 1, 0, 0]$}
  \label{tab:table1}
  \resizebox{\textwidth}{!}{
  \begin{tabular}{l||l|l|l|l|l|l|l|l|l|l|l|l|l|l|l|l|l|l|l|l||l} 
      \textbf{Steps} & \textbf{$x_0$} & \textbf{$x_1$} & \textbf{$x_2$} & \textbf{$x_3$} & \textbf{$x_4$} & \textbf{$x_5$} & \textbf{$x_6$} & \textbf{$x_7$} & \textbf{$x_8$} & \textbf{$x_9$} & \textbf{$x_{10}$} & \textbf{$x_{11}$} &\textbf{$x_{12}$} & \textbf{$x_{13}$} & \textbf{$x_{14}$} & \textbf{$x_{15}$} & \textbf{$x_{16}$} & \textbf{$x_{17}$} & \textbf{$x_{18}$} & \textbf{$\exists x'$} \\
      \hline
      1 & ? & 1 & 0 & 1 & 1 & 0 & 1 & 1 & 1 & 0 & 0 & 0 & 0 & 0 & 0 & 1 & 1 & 0 & 0 & $\bot$ \\
      2 & ? & ? & 0 & 1 & 1 & 0 & 1 & 1 & 1 & 0 & 0 & 0 & 0 & 0 & 0 & 1 & 1 & 0 & 0 & $\bot$ \\
      3 & ? & ? & ? & 1 & 1 & 0 & 1 & 1 & 1 & 0 & 0 & 0 & 0 & 0 & 0 & 1 & 1 & 0 & 0 & $\bot$ \\
      4 & ? & ? & ? & ? & 1 & 0 & 1 & 1 & 1 & 0 & 0 & 0 & 0 & 0 & 0 & 1 & 1 & 0 & 0 & $\top$ \\
      5 & ? & ? & ? & 1 & ? & 0 & 1 & 1 & 1 & 0 & 0 & 0 & 0 & 0 & 0 & 1 & 1 & 0 & 0 & $\bot$ \\
      6 & ? & ? & ? & 1 & ? & ? & 1 & 1 & 1 & 0 & 0 & 0 & 0 & 0 & 0 & 1 & 1 & 0 & 0 & $\bot$ \\
      7 & ? & ? & ? & 1 & ? & ? & ? & 1 & 1 & 0 & 0 & 0 & 0 & 0 & 0 & 1 & 1 & 0 & 0 & $\bot$ \\
      8 & ? & ? & ? & 1 & ? & ? & ? & ? & 1 & 0 & 0 & 0 & 0 & 0 & 0 & 1 & 1 & 0 & 0 & $\top$ \\
      9 & ? & ? & ? & 1 & ? & ? & ? & 1 & ? & 0 & 0 & 0 & 0 & 0 & 0 & 1 & 1 & 0 & 0 & $\bot$ \\
      10& ? & ? & ? & 1 & ? & ? & ? & 1 & ? & ? & 0 & 0 & 0 & 0 & 0 & 1 & 1 & 0 & 0 & $\top$ \\
      11& ? & ? & ? & 1 & ? & ? & ? & 1 & ? & 0 & ? & 0 & 0 & 0 & 0 & 1 & 1 & 0 & 0 & $\top$ \\
      12& ? & ? & ? & 1 & ? & ? & ? & 1 & ? & 0 & 0 & ? & 0 & 0 & 0 & 1 & 1 & 0 & 0 & $\bot$ \\
      13& ? & ? & ? & 1 & ? & ? & ? & 1 & ? & 0 & 0 & ? & ? & 0 & 0 & 1 & 1 & 0 & 0 & $\top$ \\
      14& ? & ? & ? & 1 & ? & ? & ? & 1 & ? & 0 & 0 & ? & 0 & ? & 0 & 1 & 1 & 0 & 0 & $\bot$ \\
      15& ? & ? & ? & 1 & ? & ? & ? & 1 & ? & 0 & 0 & ? & 0 & ? & ? & 1 & 1 & 0 & 0 & $\bot$ \\
      16& ? & ? & ? & 1 & ? & ? & ? & 1 & ? & 0 & 0 & ? & 0 & ? & ? & ? & 1 & 0 & 0 & $\top$ \\
      17& ? & ? & ? & 1 & ? & ? & ? & 1 & ? & 0 & 0 & ? & 0 & ? & ? & 1 & ? & 0 & 0 & $\top$ \\
      18& ? & ? & ? & 1 & ? & ? & ? & 1 & ? & 0 & 0 & ? & 0 & ? & ? & 1 & 1 & ? & 0 & $\bot$ \\
      19& ? & ? & ? & 1 & ? & ? & ? & 1 & ? & 0 & 0 & ? & 0 & ? & ? & 1 & 1 & ? & ? & $\top$ \\
  final & ? & ? & ? & 1 & ? & ? & ? & 1 & ? & 0 & 0 & ? & 0 & ? & ? & 1 & 1 & ? & 0 & \\
    \end{tabular}
  }
\end{table*}
Algorithm~\ref{algo::CEalgorithm} is illustrated on Table~\ref{tab:table1}. The explanation process follows the steps outlined in table~\ref{tab:table1}, which iteratively checks the necessity of each feature with respect to the decision.

In the first step (first row), the algorithm verifies whether feature \( x_0 \) is necessary by checking whether there exists an alternative input \( x' \), identical to $x$ except for the value of \( x_0 \), such that the decision remains unchanged. Since the check is negative (denoted by \( \bot \) in the column \( \exists x' \)), this indicates that \( x_0 \) is not necessary for the decision. As a result, its value is permanently removed for the remainder of the computation.

In the second step, the algorithm evaluates whether \( x_1 \) is necessary and again obtains a negative result (\( \bot \)). Hence, \( x_1 \) is also removed.

In contrast, Step 4 checks whether \( x_3 \) is necessary to reach the decision, and this time the result is positive (denoted by \( \top \) in the column \( \exists x' \)), indicating that \( x_3 \) is necessary. Consequently, its value is reintroduced in Step 5.

The algorithm proceeds similarly for the remaining features, testing each one to determine whether it is necessary for preserving the decision outcome, and retaining only those features that are found to be necessary.

The final row contains the minimal explanation $XP$.
\[
    XP = \{x_3 \land x_7 \land \neg x_9 \land \neg x_{10} \land \neg x_{12} \land x_{15} \land x_{16} \land \neg x_{18}\}.
\]
This means that the specific values of the features in $XP$ are sufficient to entail the outcome. Consequently, any other individual who shares these values on the features indexed by $XP$ will also be predicted to seek treatment. To express $XP$ in human-understandable terms, refer to Appendix~\ref{Apx::B1}.
\section{Methodology and Results}
\subsection{Bias assessment in the model}
As mentioned previously, $AXPs$ can also be employed to assess whether an individual decision exhibits bias. Following the definition in \cite{Darwiche2020}, a decision is considered \textbf{biased} if all its abductive explanations necessarily include a sensitive or protected attribute. In addition, the model is considered as biased if it admits at least one biased decision. We apply this notion to evaluate bias in our trained model.

More formally, the algorithm to make that check is to constraint the algorithm \ref{algo::CEalgorithm} to generate an explanation without the protected feature.
If such an explanation exists, the decision is considered unbiased with respect to that feature. Otherwise, the decision is deemed biased.

In the context of the \textit{Mental Health in Tech Survey} dataset, the feature \textit{Gender} is treated as a protected attribute, encoded as $1$ for Male and $0$ for Non-Male (Female or Trans). Accordingly, for each individual in the dataset, we examine whether there exists at least one $AXP$ that does not involve the \textit{Gender} feature. If no such explanation exists,meaning that all minimal sufficient explanations include \textit{Gender}: \textbf{the decision is deemed biased}.
\begin{table}[!ht]
  \begin{center}
    \caption{Individual bias in the decision making process}
    \label{tab:table3}
    \begin{tabular}{l|l|l} 
      \textbf{Unbiased decisions} & \textbf{Negative} & \textbf{Positive} \\
      \hline
      864& 290 & 103 \\

    \end{tabular}
  \end{center}
\end{table}

Table~\ref{tab:table3} indicates that although the model exhibits bias since it produces at least one biased decision, the majority of individual decisions (\textbf{68.73\%}) are in fact unbiased. This result highlights another strength of the abductive approach: it enables precise identification of the specific individuals affected by bias, thereby facilitating targeted mitigation strategies. Another notable finding is that the feature \textit{Gender} leads to a greater number of biased decisions associated with negative outcomes (290) compared to positive ones (103).
\begin{example}
   The individual at index $1$, who is predicted \emph{not} to seek treatment, constitutes a biased decision under our framework, as there does not exist an abductive explanation for this prediction that excludes the protected feature \textit{Gender}.
\end{example}

\subsection{Impact of features on the model output}
In this section, we examine the impact of each feature on the model’s predictions by extending our earlier analysis of the sensitive attribute \textit{Gender}. For each feature, we check whether there exists at least one abductive explanation for each decision that excludes it. If no such explanation exists, we deem the feature \emph{critical} for that prediction. A comprehensive summary of the impact of each feature is provided in table~\ref{tab:table2}.
\begin{table}[h]
  \begin{center}
    \caption{Influence of features on the model outputs}
    \label{tab:table2}
    \begin{tabular}{l|l|l|l} 
      \textbf{Features} & \textbf{Non influenced} & \textbf{Negative} & \textbf{Positive} \\
      \hline
      $x_0$&1027& 113 & 117 \\
      $x_1$&864& 290 & 103 \\
      $x_2$&1065& 107 & 85 \\
      $x_3$&630& 422 & 205 \\
      $x_4$&1126& 59 & 72 \\
      $x_5$&1109& 81 & 67 \\
      $x_6$&1122& 65 & 70 \\
      $x_7$&919& 268 & 70 \\ 
      $x_8$&1075& 76 & 106 \\
      $x_{9}$&1038& 72 & 147 \\
      $x_{10}$&1017& 71 & 169 \\
     $x_{11}$&910& 217 & 130 \\
      $x_{12}$&1032& 117 & 108 \\
      $x_{13}$&1019& 172 & 66 \\
      $x_{14}$&964& 221 & 72 \\
      $x_{15}$&888& 92 & 277 \\
      $x_{16}$&1026& 151 & 80 \\
      $x_{17}$&1138& 56 & 63 \\
      $x_{18}$&896& 273 & 88 \\
    \end{tabular}
  \end{center}
\end{table}

It emerges that the most critical feature contributing to predictions of individuals \emph{not} seeking treatment is feature $x_3$, which is indispensable in \textbf{422} decisions. This accounts for approximately $34\%$ of all cases and specifically $69.1\%$ of the instances predicted as not seeking treatment. Conversely, the most critical feature influencing predictions of individuals seeking treatment is feature $x_{15}$, appearing as critical in \textbf{277} decisions, or about $22\%$ of all cases and $42.8\%$ of the instances predicted as having sought treatment.

\subsection{The most critical combinations that impact model's outputs}
In Table~\ref{tab::critical}, we have seected the five most critical features for each category of model outputs and highlight the most prominent feature combinations associated with each outcome. Notably, the combination $(x_1 \land x_{15})$ appears in all abductive explanations of \textbf{105} instances, accounting for $16.2\%$ of the cases predicted as seeking treatment and $8.3\%$ in the whole dataset.

For the category of individuals predicted \emph{not} to seek treatment, the combination $(x_3 \land x_{11} \land x_{18})$ appears in \textbf{134} instances, representing $21.9\%$ of such predictions. Additional critical combinations are detailed in Table~\ref{tab::critical}.
\begin{table}[h]
\begin{center}
\caption{Critical combinations of features}
\label{tab::critical}
\begin{tabular}{llcc}
\toprule
\multirow{2}{*}{Outcomes} & \multirow{2}{*}{Combinations} & \multicolumn{2}{c}{Ratio} \\
\cmidrule(lr){3-4}
 & & Whole & Specific \\
\midrule
\multirow{2}{*}{Negative} 
 & $ x_3 \land x_{14} \land x_{18}$ & 9.3\% & 19.1\% \\
 & $ x_3 \land x_{11} \land x_{18}$ & 10.6\% & 21.9\% \\
 & $x_1 \land x_{14} \land x_{18}$ & 8.5\% & 17.5\% \\
 & $ x_1 \land x_3 \land x_{18}$ & 10.5\% & 21.8\% \\
 & $ x_1 \land x_3 \land x_{14}$ & 8.9\% & 18.5\% \\
\midrule
\multirow{2}{*}{Positive}
 & $ x_3 \land x_{15}$ & 8.3\% & 16.23\% \\
 & $ x_9 \land x_{15}$ & 7\%  & 13.6\%  \\
 & $ z_9 \land x_{10}$ & 7\% & 13.6\%  \\
\bottomrule
\end{tabular}
\end{center}
\end{table}

\section{Guiding Model Selection and Recourse}
By aggregating abductive explanations across individuals and comparing them to established clinical insights \citep{Stefanis2024Trends, Ali2024Ethics}, we can formally evaluate which models best reflect expected psychiatric patterns. This facilitates selecting the most appropriate model for specific mental health applications such as stress versus depression and guides recourse strategies by ensuring interventions target the true factors driving predictions.
\section*{Conclusion}
This work introduces a formal abductive reasoning framework to interpret and audit mental health-related decisions in tech workplaces. By computing abductive explanations for each model prediction, we identify the critical features and combinations of factors that drive help-seeking behavior. This not only enables clinicians and practitioners to select models that align more closely with psychiatric understanding, but also supports the design of meaningful recourse strategies.

Importantly, our method exposes potential bias in the decision processes by revealing the predictions that are sensitive to protected attributes. While some of this bias may reflect real-world asymmetries in mental health (e.g., gender-specific vulnerabilities like post-partum depression), our findings call for careful scrutiny to distinguish beneficial personalization from harmful discrimination.

Our results have significant implications for shaping workplace mental health policies and advancing Responsible AI practices. By promoting transparency and trust, we pave the way for ethical, explainable, and fairer AI-driven systems in high-stakes domains.

This study is limited to binary features; extending the approach to continuous or mixed-type data remains a compelling direction for future work. Additionally, assessing how fairness metrics evolve under abductive explanations could further deepen our understanding of model accountability.

Mental health challenges affect people of all ages, genders, and geographies. It is therefore vital to develop reliable and interpretable AI systems that empower individuals, inform policy, and ultimately foster more inclusive and supportive work environments.

\bibliographystyle{named}
\bibliography{ijcai24}
\appendix
\section{Dataset and Model}\label{Apx::A}
\subsection{Dataset processing}\label{Apx::A1}
This section lists the features used from the \textit{Mental Health in Tech Survey} dataset and describes how each was binarized.
\begin{itemize}
    \item feature $x_0$: Is the applicant older than 31? $1$ if yes, $0$ otherwise.
    \item feature $x_1$: Is the applicant male? $1$ if yes, $0$ otherwise.
    \item feature $x_2$: Is the applicant self-employed?
    \item feature $x_3$: Family history of mental health issues?
    \item feature $x_4$: Works with a small number of people?
    \item feature $x_5$: Works remotely?
    \item feature $x_6$: Works in a tech company?
    \item feature $x_7$: Aware of provided benefits?
    \item feature $x_8$: Aware of care options?
    \item feature $x_9$: Aware of employee wellness programs?
    \item feature $x_{10}$: Knows how to seek help?
    \item feature $x_{11}$: Is anonymity protected if using mental health resources?
    \item feature $x_{12}$: Is it easy to take medical leave for mental health?
    \item feature $x_{13}$: Believes discussing mental health with employer has negative consequences?
    \item feature $x_{14}$: Believes discussing physical health with employer has negative consequences?
    \item feature $x_{15}$: Comfortable discussing mental health with coworkers?
    \item feature $x_{16}$: Comfortable discussing mental health with supervisors?
    \item feature $x_{17}$: Believes employer treats mental health as seriously as physical health?
    \item feature $x_{18}$: Has observed negative consequences for coworkers with mental health conditions?
\end{itemize}
\subsection{Model training and conversion}\label{Apx::A2}

\begin{figure}[h]
\centering
\includegraphics[width=0.5\textwidth]{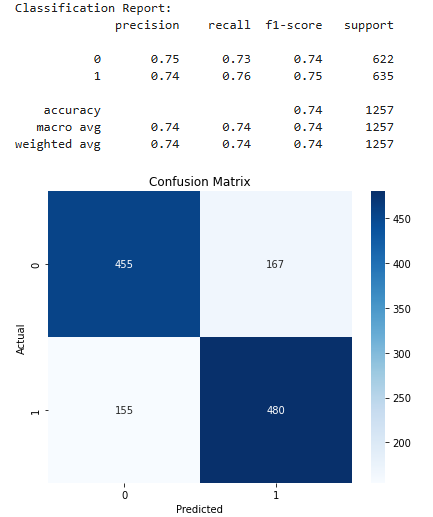}
\caption{
  Classification report of the trained model.  
}
\label{fig:CR}
\end{figure}

The decision process is implemented as a single-layer neural network, whose properties and performance are summarized in Figure~\ref{fig:CR}.

The trained neural network is then converted into an SMT formula, which serves as the basis for computing abductive explanations via constraint programming.

\section{Formal abductive explanations}\label{Apx::B}
\subsection{Computation}\label{Apx::B1}
\[
    XP = \{x_3 \land x_7 \land \neg x_9 \land \neg x_{10} \land \neg x_{12} \land x_{15} \land x_{16} \land \neg x_{18}\}.
\]
This means individual $5$ is predicted to seek treatment because:
\begin{itemize}
    \item $x_{3} = 1$: has a family history of mental health issues,
    \item $x_{7} = 1$: knows the benefits provided,
    \item $x_{9} = 0$: does not know about the wellness program,
    \item $x_{10} = 0$: does not know how to seek help at the workplace,
    \item $x_{12} = 0$: it is not easy to take leave for mental health conditions,
    \item $x_{15} = 1$: could discuss mental health with some coworkers,
    \item $x_{16} = 1$: could discuss mental health with a supervisor,
    \item $x_{18} = 0$: has not observed negative consequences for coworkers with mental health conditions.
\end{itemize}

\end{document}